\documentclass[sigconf]{acmart}
\usepackage{subcaption}
\usepackage{xspace}
\usepackage{makecell}
\usepackage{multirow}
\usepackage{amsfonts}
\usepackage{url}
\usepackage{graphicx}
\usepackage{textcomp}
\usepackage{xcolor}
\usepackage{bm}
\usepackage{algorithm,algorithmic}
\usepackage{enumitem}
\AtBeginDocument{%
  \providecommand\BibTeX{{%
    \normalfont B\kern-0.5em{\scshape i\kern-0.25em b}\kern-0.8em\TeX}}}

\copyrightyear{2025}
\acmYear{2025}
\setcopyright{cc}
\setcctype{by-nc}
\acmConference[BCB '25]{Proceedings of the 16th ACM International Conference on Bioinformatics, Computational Biology, and Health Informatics}{October 11--15, 2025}{Philadelphia, PA, USA}
\acmBooktitle{Proceedings of the 16th ACM International Conference on Bioinformatics, Computational Biology, and Health Informatics (BCB '25), October 11--15, 2025, Philadelphia, PA, USA}
\acmDOI{10.1145/3765612.3767246}
\acmISBN{979-8-4007-2200-4/2025/10}

\begin{document}

\title{
Developing Fairness-Aware Task Decomposition to Improve Equity in Post-Spinal Fusion Complication Prediction
}

\author{Yining Yuan}
\authornote{Both authors contributed equally to this research.}
\affiliation{%
  \institution{Georgia Institute of Technology}
  \city{Atlanta}
  \state{GA}
  \country{USA}
}
\email{yyuan394@gatech.edu}

\author{J. Ben Tamo }
\authornotemark[1]
\affiliation{%
  \institution{Georgia Institute of Technology}
  \city{Atlanta}
  \state{GA}
  \country{USA}
}
\email{jtamo3@gatech.edu}

\author{Wenqi Shi}
\affiliation{%
  \institution{UT Southwestern Medical Center}
  \city{Dallas}
  \state{Texas}
  \country{USA}
}
\email{wenqi.shi@utsouthwestern.edu}

\author{Yishan Zhong}
\affiliation{%
  \institution{Georgia Institute of Technology}
  \city{Atlanta}
  \state{GA}
  \country{USA}
}
\email{yzhong307@gatech.edu}

\author{Micky C. Nnamdi}
\affiliation{%
  \institution{Georgia Institute of Technology}
  \city{Atlanta}
  \state{GA}
  \country{USA}
}
\email{mnnamdi3@gatech.edu}

\author{B. Randall Brenn}
\affiliation{%
  \institution{Shriners Children’s Hospital}
  \city{Philadelphia}
  \state{PA}
  \country{USA}
}
\email{bbrenn@shrinenet.org}

\author{Steven W. Hwang}
\affiliation{%
  \institution{Shriners Children’s Hospital}
  \city{Philadelphia}
  \state{PA}
  \country{USA}
}
\email{sthwang@shrinenet.org}

\author{May D. Wang}
\authornote{Corresponding author.}
\affiliation{%
  \institution{Georgia Institute of Technology}
  \city{Atlanta}
  \state{GA}
  \country{USA}
}
\email{maywang@gatech.edu}

\renewcommand{\shortauthors}{Yuan et al.}
\begin{abstract}
Fairness in clinical prediction models remains an open challenge, as many methods oversimplify outcomes and inadvertently propagate demographic biases. This issue is especially consequential in spinal fusion surgery for scoliosis, a high-risk procedure with heterogeneous patient outcomes.
We present FAIR-MTL, a fairness-aware multitask learning framework for equitable and fine-grained prediction of postoperative complication severity.
Instead of relying on explicit sensitive attributes during model training, FAIR-MTL employs a data-driven subgroup inference mechanism inspired by Sensitive Set Invariance (SSI)\cite{yurochkin2021sensitive}. We extract a compact demographic embedding—capturing features such as age and gender—and apply k-means clustering to uncover latent patient subgroups that may be differentially affected by traditional models. These inferred subgroup labels determine task routing within a shared multitask architecture: a common encoder learns global patient representations, while task-specific heads specialize in modeling patterns unique to each subgroup. During training, subgroup imbalance is mitigated through inverse-frequency weighting, and regularization prevents overfitting to smaller groups.
FAIR-MTL achieves an AUC of 0.86 and accuracy of 75\% across four complication severity classes while reducing the average demographic parity difference to 0.055 and equalized odds difference to 0.094 for gender, and 0.056 and 0.148, respectively for age, substantially improving fairness over standard baselines. SHAP and Gini importance analyses highlight clinically relevant predictors, including hematocrit, hemoglobin, and patient weight, providing transparency at both global and individual levels. Our results demonstrate that integrating fairness-aware task decomposition into model design enables equitable, interpretable, and clinically actionable predictions for surgical risk stratification.

\end{abstract}




\maketitle

\section{Introduction}

\begin{figure*}[t]

    \centering
    \includegraphics[width=0.93\textwidth]{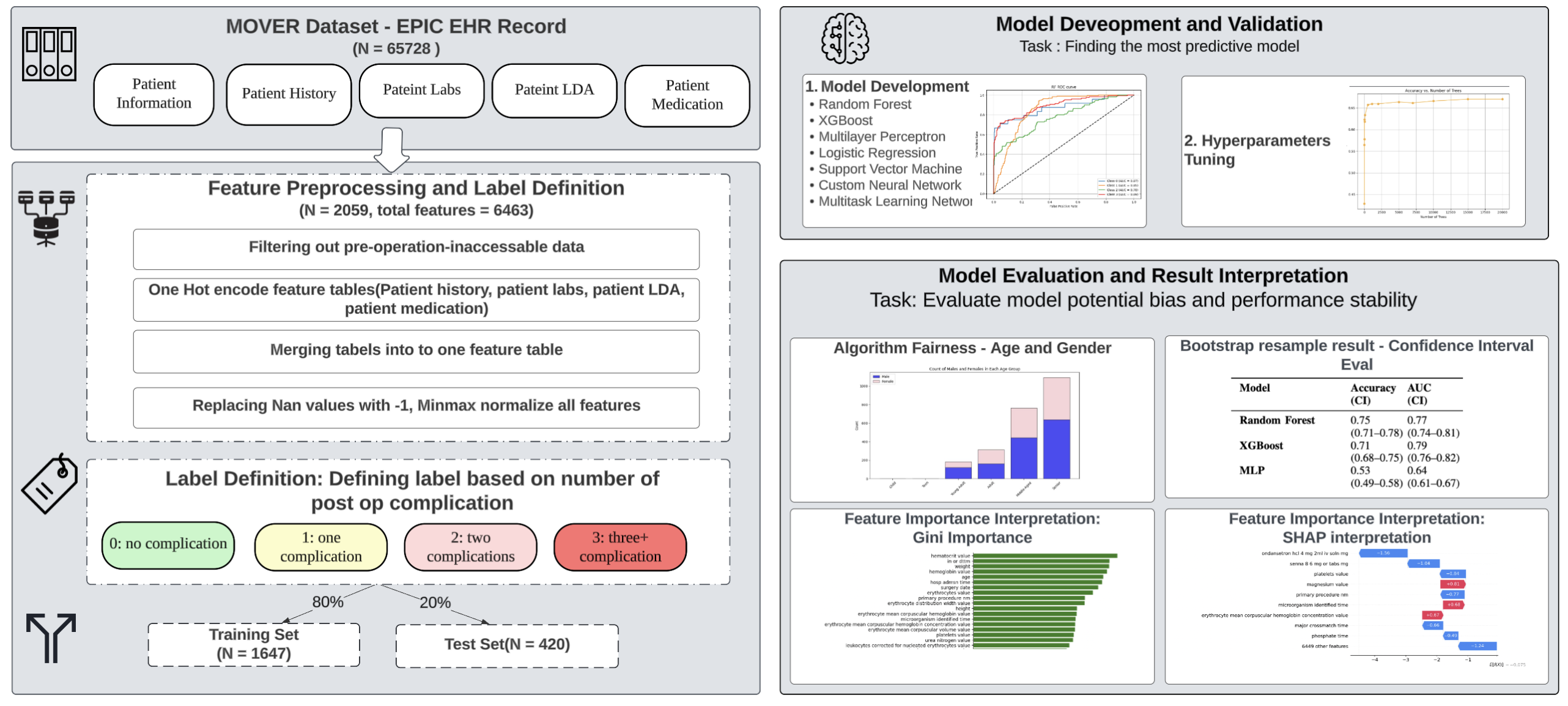}
    
    \caption{
    Overview of the proposed responsible AI framework for predicting postoperative complication severity, enhancing personalized surgery decision-making.
    1. Feature Preprocessing and Label Definition: Pre-operative data, including patient history, lab results, medication, Lines/Drains/Airways, and patient basic information are processed to create a unified feature table. 
    2. Model Development and Validation: Five different machine learning models are developed and tuned using grid search to optimize hyperparameters. 
    3. Model Evaluation and Result Interpretation: The trained models are evaluated for fairness across age and gender, performance stability through bootstrap resampling, and feature importance using Gini importance and SHAP interpretations. 
    }

    \label{fig:overview}
\end{figure*}

Spinal surgery is a complex field of medicine where clinical decisions significantly impact patient outcomes. One of the most challenging and prevalent procedures within spinal surgery is spinal fusion, a surgical intervention performed to stabilize the vertebral column in cases of spinal instability, deformities, or degenerative disc diseases \cite{deyo2004spinal}. Each year, more than 450,000 spinal fusion procedures are performed in the United States alone \cite{fingar2015most, goz2013perioperative}, highlighting its widespread use and clinical importance.
Recent studies have further demonstrated heterogeneous treatment effects in adolescent idiopathic scoliosis patients \cite{tamo2024heterogeneous} and introduced predictive modeling approaches to anticipate patient-reported outcomes, highlighting the need to integrate computational perspectives into clinical decision-making \cite{nnamdi2023concept}.

Despite its frequency, spinal fusion carries a high risk of complications, ranging from minor infections to severe events such as respiratory failure or neurological deficits, leading to prolonged recovery, higher costs, and reduced quality of life \cite{tang2014risk, kim2014operative}. These complications, which range from minor issues such as wound infections to severe events like respiratory failure or neurological deficits, pose substantial risks to patients.
However, current surgical risk assessment tools are often constrained by their reliance on static patient characteristics and generalized risk scores \cite{willems2013decision,  eamer2018review}, which fail to capture the nonlinear relationships and complex interactions influencing surgical outcomes. Additionally, many existing models in both clinical and machine‑learning applications focus primarily on improving robustness and predictive performance, with limited attention to fairness \cite{feng2022fair}. This lack of fairness consideration can inadvertently perpetuate or even amplify existing healthcare disparities.

Although the broader field of fairness-aware machine learning has made notable methodological advances, introducing techniques such as adversarial debiasing, sample reweighting, and fairness-driven regularization to mitigate biases across predefined sensitive attributes like race, gender, or age \cite{mehrabi2021survey, pfohl2021empirical, zhang2020fairness}, their application to clinical prediction tasks has been sparse. In cases where fairness interventions are applied, they are often implemented in isolated stages (pre-processing, in-processing, or post-processing), rather than embedded into a unified, end-to-end framework.
Furthermore, many fairness-aware algorithms prioritize reducing disparities at the expense of predictive accuracy, offering limited strategies for balancing both goals in high-stakes clinical decision-making.

The main contributions of our work are:
\begin{itemize}
\item \textbf{Fairness-Aware Modeling:} We propose a fairness-aware multi-task learning framework that leverages Sensitive Subgroup Inference to detect latent disparity-related patient subgroups and integrates them through task routing for subgroup-specific prediction while maintaining shared learning across populations. 
\item \textbf{Transparency and Interpretability: } We apply explainable AI techniques, including SHAP and feature importance analysis, to provide interpretability at both the global and patient levels.. These explanations help clinicians understand the model’s rationale, increasing trust and aiding in shared decision-making.
\end{itemize}

\section{Related Work}

\subsection{AI in Predictive Modeling for Postoperative Outcomes in Spine Surgery}
The application of machine learning (ML) to predict postoperative outcomes in spine surgery has advanced quickly \cite{ben2023adolescent, shi2025predicting}, opening new opportunities for risk stratification and individualized care. For example, Schonnagel et al.\ applied several ML approaches to forecast persistent lower back pain two years after lumbar spondylolisthesis surgery, finding that gradient-boosted models achieved strong predictive performance with an AUC of 0.81 \cite{schonnagel2024predicting}. Similarly, Khor et al.\ used a statewide dataset to estimate one-year outcomes, including functional disability and pain reduction. Logistic regression achieved moderate accuracy, with an AUC of 0.79 for predicting back pain improvement \cite{khor2018development}. Shah et al.\ also applied gradient boosting to identify patients at risk of major complications or readmission following lumbar fusion, supplementing their predictions with visualizations that clarified how different clinical features influenced outcomes \cite{shah2021prediction}. At a broader level, Hassan et al.\ emphasized that ML consistently outperforms traditional surgical risk calculators, while White et al.\ highlighted its potential to support more patient-centered decision-making \cite{hassan2023artificial,white2020predicting}.  

Although these studies demonstrate clear predictive value, most have not addressed fairness considerations, raising concerns that performance may vary across demographic groups and potentially exacerbate existing disparities.  

\subsection{Ethics and Fairness in Clinical AI}
Healthcare AI poses unique ethical challenges due to the complexity of medical data, societal implications, and clinical stakes. Naik et al. \cite{naik2022legal} emphasize the legal and ethical tensions between AI utility and privacy. Morley et al. categorize challenges into three areas: epistemic (e.g., probabilistic, opaque, or biased evidence), normative (e.g., fairness, autonomy), and traceability (e.g., black-box accountability) \cite{morley2020ethics}. 

Critically, algorithms trained on real-world clinical data often reflect historical and systemic biases. For example, Obermeyer et al. showed that a widely deployed commercial algorithm significantly underestimated the care needs of Black patients, allocating fewer resources for comparable health profiles \cite{obermeyer2019dissecting}. Kordzadeh and Agarwal further documented how algorithmic decision systems may reinforce structural inequalities in healthcare delivery \cite{kordzadeh2022algorithmic, agarwal2023addressing}.

To improve transparency, methods such as SHapley Additive exPlanations (SHAP) have become prominent. SHAP is a model-agnostic technique that computes feature importance by marginalizing all possible feature coalitions \cite{lundberg2020local}. Its utility has been demonstrated in various clinical applications, where both local and global explanations can support clinician trust and validation \cite{shi2025predicting, guo2024interpretable}.

\subsection{Fairness-Aware Learning and Task Decomposition in AI
}

Fairness-aware machine learning aims to reduce algorithmic disparities across sensitive attributes (e.g., race, gender, age) while maintaining performance. Techniques such as reweighting, adversarial debiasing, and fairness-constrained optimization are commonly employed, though often in isolation (pre-, in-, or post-processing) rather than within an integrated modeling framework.

An emerging alternative is to integrate fairness objectives directly into the learning architecture. Fang et al. \cite{oneto2019taking} proposed using multi-task learning (MTL) for fair classification by treating fairness constraints (e.g., equalized odds) as auxiliary prediction tasks. Their work demonstrated that MTL can balance fairness and accuracy by encouraging shared representations that align predictive and fairness goals.

Hosseini et al. introduced Group-Aware Learning, an in-processing method that jointly optimizes accuracy and fairness by penalizing disparities in false positive/negative rates across groups \cite{hosseini2023fair}. Yurochkin and Sun proposed SenSeI, which enforces invariance across sensitive sets through regularization \cite{yurochkin2021sensei}. Zhang et al. recently proposed Fair-MoE, a fairness-oriented Mixture-of-Experts model that dynamically routes inputs across expert models to balance subgroup accuracy and demographic equity \cite{wang2025fairmoe}.

However, few of these models are applied in clinical contexts, and most rely on either pre-, in-, or post-processing fairness techniques. Building on this line of research, our work introduces a fairness-aware multi-task learning approach that combines Sensitive Subgroup Inference (SSI) with task-specific routing, representing a hybrid pre- and in-processing technique. Rather than optimizing fairness metrics directly, we enable fairness-aware modeling by structuring the prediction task around subgroup-specific heads within a unified MTL framework. This design allows our model to capture both global trends and subgroup-specific risks in a high-stakes clinical setting, thereby reducing disparities in outcome prediction without explicitly enforcing fairness constraints.

\section{Method}

\begin{figure}

    \centering
    \includegraphics[width=0.9\linewidth]{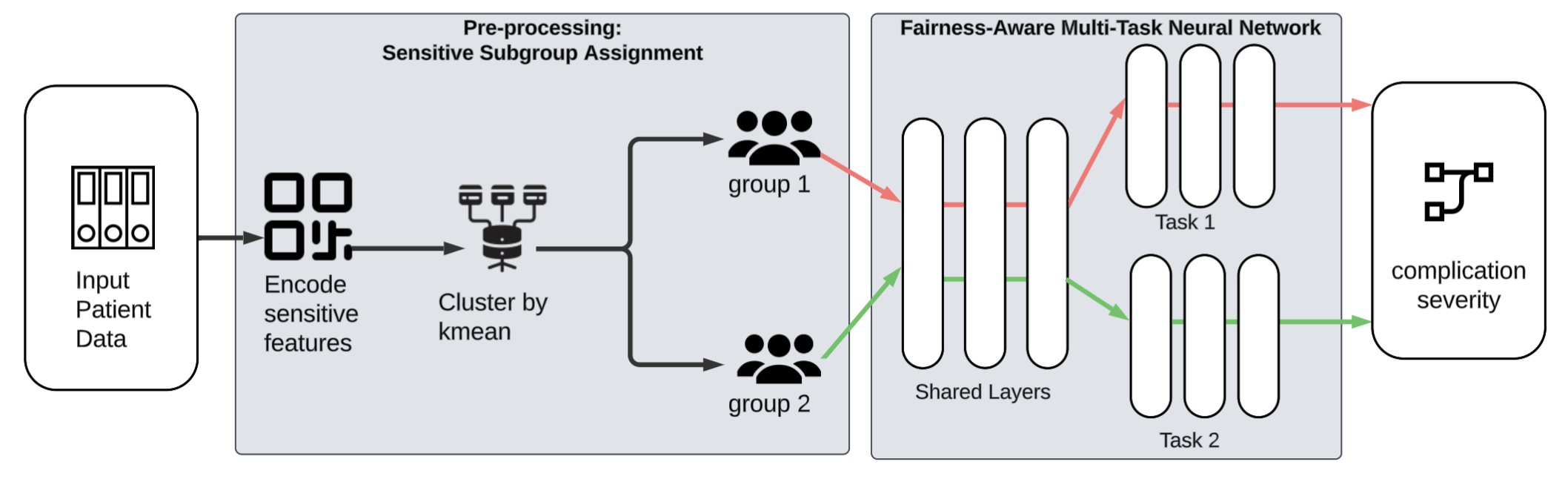}
    \caption{Overview of the proposed Fairness-Aware Multi-Task Learning Framework. Patient data are first processed to extract sensitive features, which are clustered using k-means to assign individuals into sensitive subgroups (e.g., Group 1 and Group 2). These subgroup labels are then used to route inputs through a shared neural network backbone into corresponding task-specific heads. Each task head specializes in learning patterns unique to its subgroup while sharing common representations. The final output predicts the severity of postoperative complications.}
    \label{fig:enter-label}
\end{figure}

\subsection{Problem Formulation}

Let $\mathcal{D} = \{(x_i, y_i)\}_{i=1}^N$ be a dataset of $N$ spinal fusion cases, where $x_i \in \mathbb{R}^d$ denotes the vector of preoperative features (demographics, vitals, comorbidities, etc.) and $y_i \in \{0, 1, 2, 3\}$ is a categorical variable representing the severity level of postoperative complications, ranging from no complications ($0$) to high-severity complications ($3$).

Our goal is to learn a predictive function $f: \mathbb{R}^d \rightarrow \{0, 1, 2, 3\}$ that maps each patient’s features to a probability distribution over the four outcome classes, while ensuring equitable performance across sensitive patient subgroups.

\subsection{Sensitive Subgroup Inference}

We adopt a pre-processing fairness approach inspired by Sensitive Set Invariance (SSI) \cite{yurochkin2021sensitive}, which aims to mitigate disparities by discovering latent structure within patient populations. 
We employ an unsupervised pre-processing step to identify sensitive subgroups based on two clinically relevant bias-associated variables: age and gender, both of which are known to influence surgical outcomes. The underlying assumption is that patients from certain demographic or clinical subpopulations may be at differential risk of misclassification or adverse outcomes.

Let $\mathcal{K} = \{1, \dots, K\}$ denote a partition of the dataset into $K$ clusters (subgroups), inferred from the embedding of demographic features using an unsupervised algorithm, in our case $k$-means.

Each datapoint $x_i$ is assigned a cluster label $z_i \in \{1, \dots, K\}$, such that:
\begin{equation}
z_i \;=\; \operatorname{kmeans}\!\big(\,\phi(x_i^{(d)})\,\big),
\qquad z_i \in \{1,\dots,K\}.
\end{equation}

where $x_i^{(d)}$ denotes the demographic features of patient $i$, and $\phi(\cdot)$ is an autoencoder function.

These cluster assignments are then used to route each input to a task head in the multi-task model.

\subsection{FAIR-MTL Architecture}
We model the problem as a multi-task classification problem, where each task corresponds to predicting complications for a specific SSI-derived subgroup $K$. The architecture consists of:

\begin{itemize}
    \item A shared encoder $h_s(x_i; \theta_s)$ maps input features $x_i$ into a shared latent representation. 
    \item $K$ task-specific heads $f_k(h; \theta_k)$, each parameterized by $\theta_k$, which specialize in predicting outcomes for subgroup $k$.

\end{itemize}
The model prediction is given by:
\begin{equation}
    \hat{y}_i = f_{z_i}(h_s(x_i; \theta_s); \theta_{z_i})
\end{equation}
where $h_s(\cdot)$ denotes the shared encoder and $f_{z_i}(\cdot)$ is the head for subgroup $z_i$, $\theta_s$ denotes the parameters of the shared encoder $h_s(\cdot)$, which is trained on the entire dataset to capture global structure, and $\theta_{z_i}$ represents the parameters of the task-specific head $f_{z_i}(\cdot)$, specialized for the subgroup $z_i$ to capture group-specific patterns and reduce disparity in prediction performance.

To mitigate imbalance across sensitive subgroups, we apply an inverse-frequency weighting scheme during training. Let $n_k$ denote the number of samples in subgroup $k$ and define normalized weights $w_k = \frac{1/n_k}{\sum_{j=1}^K 1/n_j}$. The final loss is a weighted categorical cross-entropy:
\begin{equation}
\mathcal{L} = - \sum_{i=1}^N w_{z_i} 
\sum_{c=0}^{C-1} \mathbf{1}_{\{y_i = c\}} 
\log \hat{y}_i^{(c)} 
\end{equation}

where $C = 4$ is the number of complication severity classes and $z_i$ is the subgroup assignment of sample $i$ via SSI.

To regularize learning and prevent overfitting to small subgroups, we apply $\ell_2$ weight regularization:

\begin{equation}
    \mathcal{L}_{\text{total}} = \mathcal{L} + \lambda \sum_{k=1}^K \|\theta_k\|^2 
\end{equation}

\subsection{Ethical Consideration}

\subsubsection{Model Explainability}

Model explainability is essential in healthcare applications to ensure trust, interpretability, and actionable insights from predictions. 

SHAP provides local interpretability by assigning each feature a contribution value for individual predictions. It is based on cooperative game theory and computes the Shapley value for each feature. The SHAP value for feature $j$ in instance $i$ is defined as:

\begin{equation}
\phi_j^{(i)} = \sum_{S \subseteq F \setminus \{j\}} \frac{|S|! (|F| - |S| - 1)!}{|F|!} \left[ f(S \cup \{j\}) - f(S) \right],
\end{equation}

where $F$ is the set of all features, $S$ is a subset of $F$ that does not include feature $j$, $f(S)$ is the model prediction when only features in $S$ are considered, and $|S|$ is the cardinality of $S$.  
For global interpretability, tree-based models provide feature importance scores that measure the overall contribution of each feature across the entire dataset. In tree-based models, feature importance is calculated as the total reduction in the impurity (e.g., Gini impurity or entropy) achieved by splits on a given feature across all trees. The feature importance for feature $j$ is expressed as:

\begin{equation}
I_j = \sum_{t \in T} \sum_{s \in S_t(j)} \Delta i(s),
\end{equation}

where $T$ is the set of all trees in the model, $S_t(j)$ is the set of splits on feature $j$ in tree $t$, and $\Delta i(s)$ is the reduction in impurity caused by split $s$. 

\subsubsection{Model Fairness}

To quantify fairness, we evaluated the model's performance across demographic groups defined by sensitive attributes such as gender and age. Fairness metrics, including demographic parity and equalized odds, were computed separately for each class label (0 to 3), reforming the problem into 4 binary classification problems of one-versus-rest. 
\textit{Demographic parity} requires that predictions are independent of the sensitive attribute $A$, i.e.,
\[
P(\hat{y} = 1 \mid A = 0) = P(\hat{y} = 1 \mid A = 1).
\]

\textit{Equalized odds} requires that predictions are conditionally independent of $A$ given the true label $y$. Formally,
\[
P(\hat{y} = 1 \mid y = 1, A = 0) = P(\hat{y} = 1 \mid y = 1, A = 1),
\]
\[
P(\hat{y} = 1 \mid y = 0, A = 0) = P(\hat{y} = 1 \mid y = 0, A = 1).
\]

These metrics were computed on both baseline models and our FAIR-MTL to evaluate the impact of our proposed model.
\section{Results}
We evaluate our model on a retrospective clinical dataset of spinal fusion patients, where the task is to predict postoperative complication severity across four ordinal classes: no, mild, moderate, or severe complications. The dataset includes demographic and clinical features, with age and gender identified as potential sources of bias. To address this, we apply clustering on bias-related variables to infer sensitive subgroups, used for task decomposition in FAIR-MTL. We compare our model against standard baselines including Extreme Gradient Boosting (XGBoost), Support Vector Machine (SVM), Random Forest (RF), and (4) Neural Network. All models are trained using stratified splits (70/15/15) and evaluated using accuracy, precision, recall, F1-score, AUROC, and fairness metrics (DP and EO) across subgroups. To assess generalizability, we also perform external validation on the INSPIRE perioperative dataset \cite{lim2024inspire,lim2024inspire1, physiobank2000physionet}, formulating postoperative mortality as a binary outcome.

\begin{table*}[]
\small
\centering
\caption{{Description of variables and their respective details, including patient information, medications, lines/drains/airway devices, labs, measurements, and postoperative complications.}}
\label{tab:dataset}
% \scalebox{0.65}{
\resizebox{1\linewidth}{!}{
\begin{tabular}{lll}
\toprule
\textbf{Variable}          & \textbf{Description} & \textbf{Detail}                      \\ 
\midrule
\textbf{Patient Information}        &                                                                    &                                         \\
\midrule
Age                     & Patient age at time of admission & $50.7 \pm 1.3$ years \\
Sex                     & Patient gender & Most frequent: Male (58.1\%) \\
Height                  & Patient height in centimeters & $170.4 \pm 10.9$ cm \\
Weight                  & Patient weight in kilograms & $82.0 \pm 20.8$ kg \\
Surgery Time            & Surgery start time & --- \\
Anesthesia Time         & Duration of anesthesia in hours & --- \\
ASA Status              & ASA physical status classification & Most frequent: Severe Systemic Disease (63.5\%) \\
Discharge Disposition   & Discharge location (e.g., home, rehab, ICU) & Most frequent: Home Routine (50.3\%) \\

\midrule
\textbf{Patient Medications} & & \\
\midrule
Medication Name        & Name of the prescribed medication & Unique Medications: 1979 \\
Medication Dose        & Dosage of the prescribed medication & Range: (min, max) not provided \\
Medication Route       & Route of administration (e.g., oral, IV) & Example: IV, Oral, Topical \\
Time of Administration & Time when the medication was administered & --- \\

\midrule
\textbf{Patient Lines, Drains, and Airway Devices (LDA)} & & \\
\midrule
Line Type             & Type of line (e.g., central venous line, arterial line) & Unique Lines/Drains: 31 \\
Drain Type            & Type of drain (e.g., chest drain, Foley catheter) & Included in Lines/Drains \\
Airway Device         & Airway device used (e.g., endotracheal tube, LMA) & Unique Airway Devices: 4 \\
Time of Placement     & Time of placement of line, drain, or device & --- \\
Time of Removal       & Time of removal of line, drain, or device & --- \\
Location of Placement & Anatomical location of line, drain, or device & Examples: Abdomen, Antecubital \\

\midrule
\textbf{Patient Labs} & & \\
\midrule
Lab Test Name       & Name of the laboratory test ordered & Unique Labs: 549 \\
Lab Result          & Observed measurement or result & --- \\
Time of Lab Result  & Time when the lab result was recorded & --- \\
Reference Range     & Reference range for lab values & Abnormal flag present \\

\midrule
\textbf{Patient Postoperative Complications} & & \\
\midrule
Complication Type        & Type of postoperative complication & Unique Types: 10 \\
Complication Description & Detailed description or free-text notes & --- \\
ICU Admission            & Whether admitted to ICU postoperatively & --- \\
Mortality                & Whether patient died postoperatively & --- \\

\midrule
\textbf{Patient Procedure Events} & & \\
\midrule
Preoperative Events  & Procedures or events before surgery & --- \\
Perioperative Events & Events during surgery & --- \\
Postoperative Events & Procedures or events after surgery & --- \\

\bottomrule
\end{tabular}
}
\end{table*}

\begin{figure}
    \centering
    \includegraphics[width=1\linewidth]{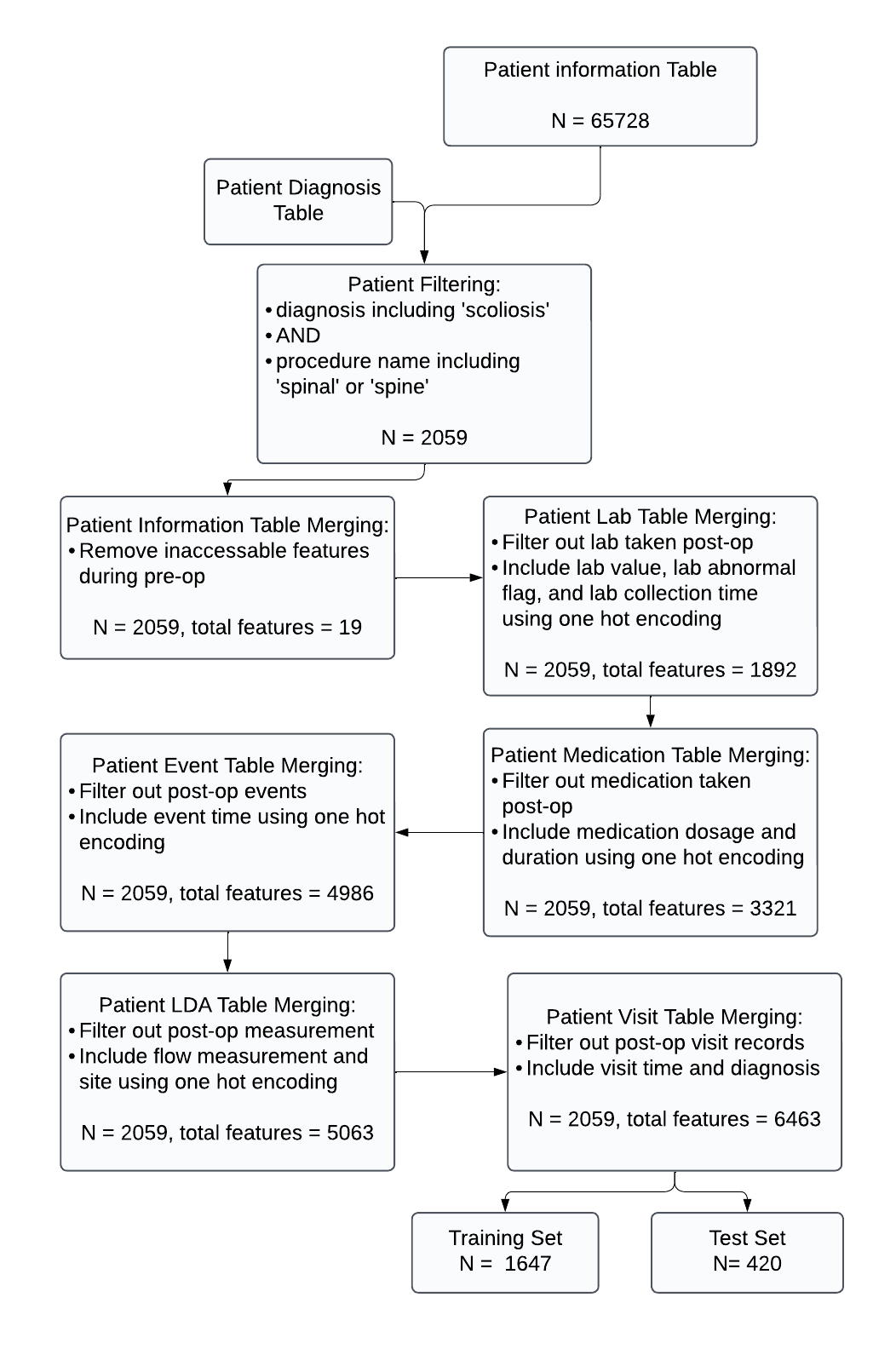}
    \caption{Data preprocessing workflow. The patient information table, initially containing 65,728 records, is filtered based on scoliosis diagnosis and spinal-related procedures, resulting in 2,059 patients. 
    % Multiple data sources, including patient information, medication, lab, event, LDA, and visit tables, are merged sequentially, applying filters such as excluding post-operative data and using one-hot encoding for relevant features. 
    The final dataset includes 6,463 features and is split into training (N = 1,647) and test (N = 420) sets.}
    \label{fig:preprocessing}
\end{figure}

\subsection{Datasets}
The dataset used for this study is sourced from the publicly available  Medical Informatics Operating Room Vitals and Events Repository (MOVER) \cite{samad2023medical}. It includes two distinct datasets derived from different electronic health record (EHR) systems: the SIS dataset and the EPIC dataset. 
The SIS dataset comprises data from 19,114 patients and their associated surgeries, collected over a two-year period from 2015 to 2017. In contrast, the EPIC dataset spans a five-year period from 2017 to 2022 and is significantly larger, containing data from 39,685 patients and a total of 65,728 surgeries.

Using the MOVER-EPIC dataset, we identified 2,059 patients diagnosed with scoliosis who underwent spinal fusion surgery, incorporating preoperative data only for making prediction. 
A comprehensive description of the dataset features is presented in Table \ref{tab:dataset}, with preprocessing steps outlined in Figure \ref{fig:preprocessing}.

To evaluate the generalization of our approach, we evaluated all baseline models and the proposed FAIR-MTL approach on a secondary dataset - INSPIRE dataset\cite{lim2024inspire}, which is a large-scale perioperative research dataset containing over 130,000 surgical cases collected from 2011 to 2020 in South Korea. The dataset is organized into relational tables that capture a wide range of perioperative information, including surgical procedure records, high-resolution intraoperative vital signs, laboratory test results, medication administration records, and postoperative outcomes. We perform preprocessing by first integrating multiple sources, including laboratory results, medications, diagnoses, and surgical operations to construct a comprehensive feature set. Feature preprocessing followed the methodology outlined Figure \ref{fig:preprocessing}, with categorical variables such as medications and laboratory tests represented using one-hot encoding. Patients were included in the analysis only if complete information was available in all relevant tables, ensuring that no feature values were missing for the selected cohort.

\subsection{Implementation Details}

We implement all models using PyTorch. The shared encoder in FAIR-MTL consists of fully connected layers with hidden dimensions $[2048, 64, 4096, 512, 2048, 128, 32]$, each followed by Batch Normalization, ReLU activation, and Dropout (rate = 0.23). Each task-specific head is a shallow MLP ending in a Softmax layer for 4-class prediction. The model is trained using the AdamW optimizer with a learning rate of $4.02 \times 10^{-5}$, batch size of 64, and a learning rate scheduler (ReduceLROnPlateau) based on validation macro-F1. Training is performed for up to 100 epochs with early stopping (patience = 10).

\begin{table}[ht]
\centering
\caption{Overall performance and subgroup accuracy metrics for predicting postoperative complication severity in spinal fusion surgery. Performance is reported using Accuracy, AUC, Precision, Recall, and F1-score. Subgroup accuracies are reported for sensitive attributes: \textbf{Sex} (Female, Male) and \textbf{Age Group} (Group 1: 0–18 years, Group 2: 19–35 years, Group 3: 36–50 years, Group 4: 51+ years). Subgroup accuracy represents the average accuracy across four complication severity classes, where smaller differences between subgroups indicate more equitable model behavior.}
\resizebox{1\linewidth}{!}{
\begin{tabular}{lccccc}
\toprule
\textbf{Metric} & \textbf{Random Forest} & \textbf{XGBoost} & \textbf{SVM} & \textbf{Neural Network} & \textbf{FAIR-MTL} \\
\midrule
\multicolumn{6}{l}{\textbf{Overall Performance}} \\
\midrule
Accuracy (CI)           & 76\% (73–80\%) & 73\% (68–78\%) & 55\% (51–60\%) & 77\% (73–82\%) & 75\% (70–79\%) \\
AUC (CI)                & 79\% (76–83\%) & 79\% (76–82\%) & 64\% (61–67\%) & 89\% (85–92\%) & 86\% (81–89\%) \\
Precision               & 75\%           & 73\%           & 46\%           & 76\%           & 74\%           \\
Recall                  & 72\%           & 69\%           & 55\%           & 71\%           & 73\%           \\
F1-score                & 70\%           & 70\%           & 44\%           & 75\%           & 72\%           \\
\specialrule{\heavyrulewidth}{0pt}{0pt} % Bold horizontal line separating sections

\multicolumn{6}{l}{\textbf{Subgroup Accuracy by Sex}} \\
\midrule
Female                  & 75.4\%         & 69.6\%         & 54.6\%         & 75.0\%         & 76.4\%         \\
Male                    & 72.8\%         & 71.1\%         & 79.29\%         & 83.2\%         & 74.5\%         \\

\midrule
\multicolumn{6}{l}{\textbf{Subgroup Accuracy by Age Group}} \\
\midrule
Group 1 (0–18)          & 69.8\%         & 67.4\%         & 51.2\%         & 79.1\%         & 75.0\%         \\
Group 2 (19–35)         & 78.8\%         & 67.5\%         & 58.8\%         & 85.0\%         & 76.5\%         \\
Group 3 (36–50)         & 72.7\%         & 67.2\%         & 49.2\%         & 75.4\%         & 74.2\%         \\
Group 4 (51+)           & 75.0\%         & 73.8\%         & 51.6\%         & 76.3\%         & 74.4\%         \\

\bottomrule
\end{tabular}}
\label{tab:performance_metrics}
\end{table}

\begin{figure}[ht]
    \centering    \includegraphics[width=0.49\linewidth]{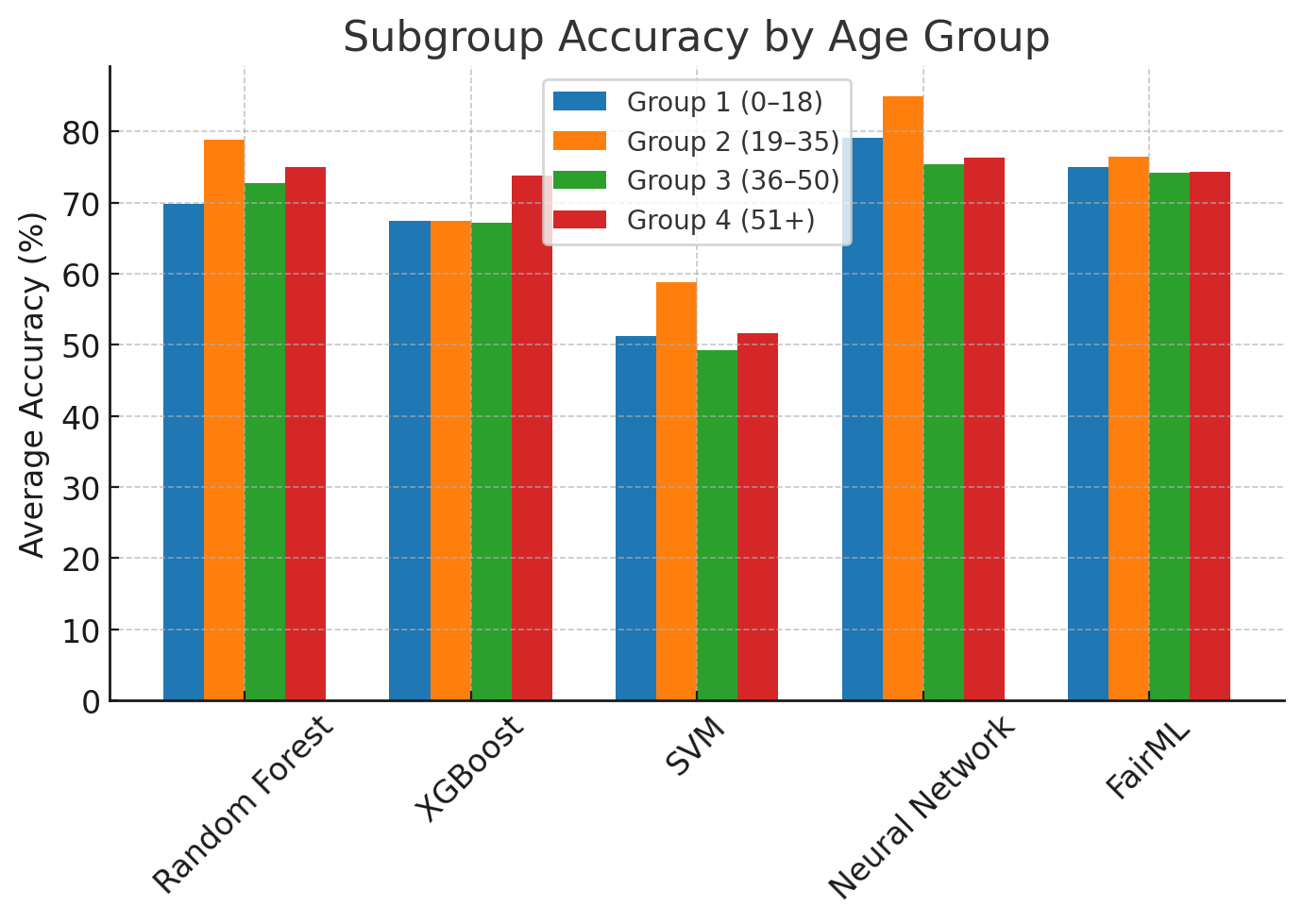}
    \includegraphics[width=0.49\linewidth]{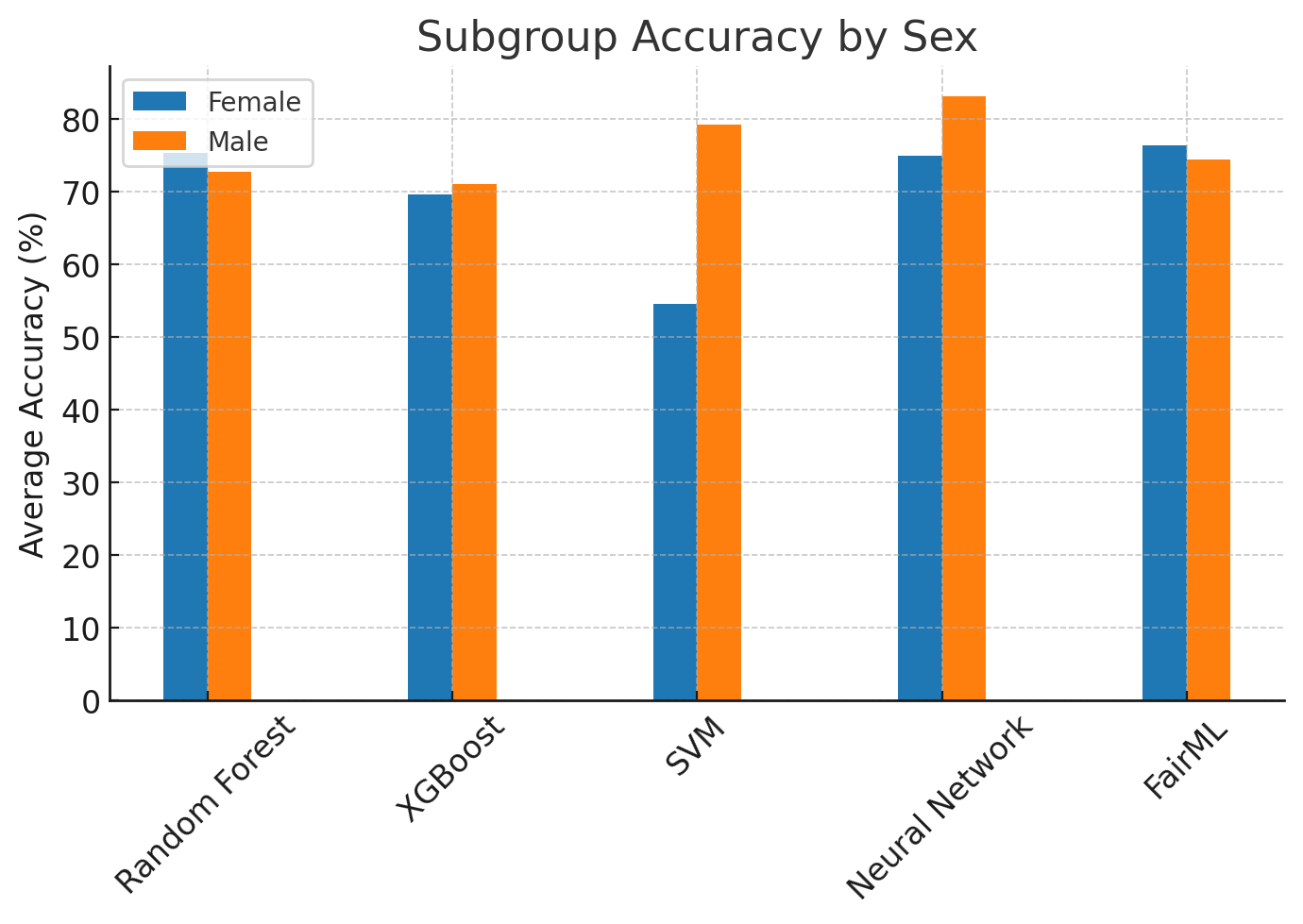}
    \caption{Subgroup accuracy comparison across machine learning models for predicting postoperative complication severity. FAIR-MTL shows more consistent accuracies across subgroups.}
    \label{fig:subgroup_accuracy}
\end{figure}

\begin{table}[ht]
\centering

\caption{Fairness summary metrics for each model. DP = Demographic Parity Difference, EO = Equalized Odds Difference. Lower values indicate better fairness on MOVER dataset.}
\resizebox{0.85\linewidth}{!}{
\begin{tabular}{lcccc}
\toprule
\textbf{Model} & \textbf{DP (Gender)} & \textbf{DP (Age)} & \textbf{EO (Gender)} & \textbf{EO (Age)} \\
\midrule
RF        & 0.065  & 0.101  & 0.087  & 0.380  \\
NN        & 0.066  & 0.145  & 0.138  & 0.444  \\
FAIR-MTL  & 0.055  & 0.056  & 0.094  & 0.148  \\
SVM  & 0.079  & 0.194  & 0.191  & 0.314  \\
XGB  & 0.075  & 0.112  & 0.140  & 0.349  \\ 

\bottomrule
\end{tabular}
}
\label{tab:fairness_metrics_combined}
\end{table}

\begin{figure}

    \centering
    \includegraphics[width=0.7\linewidth]{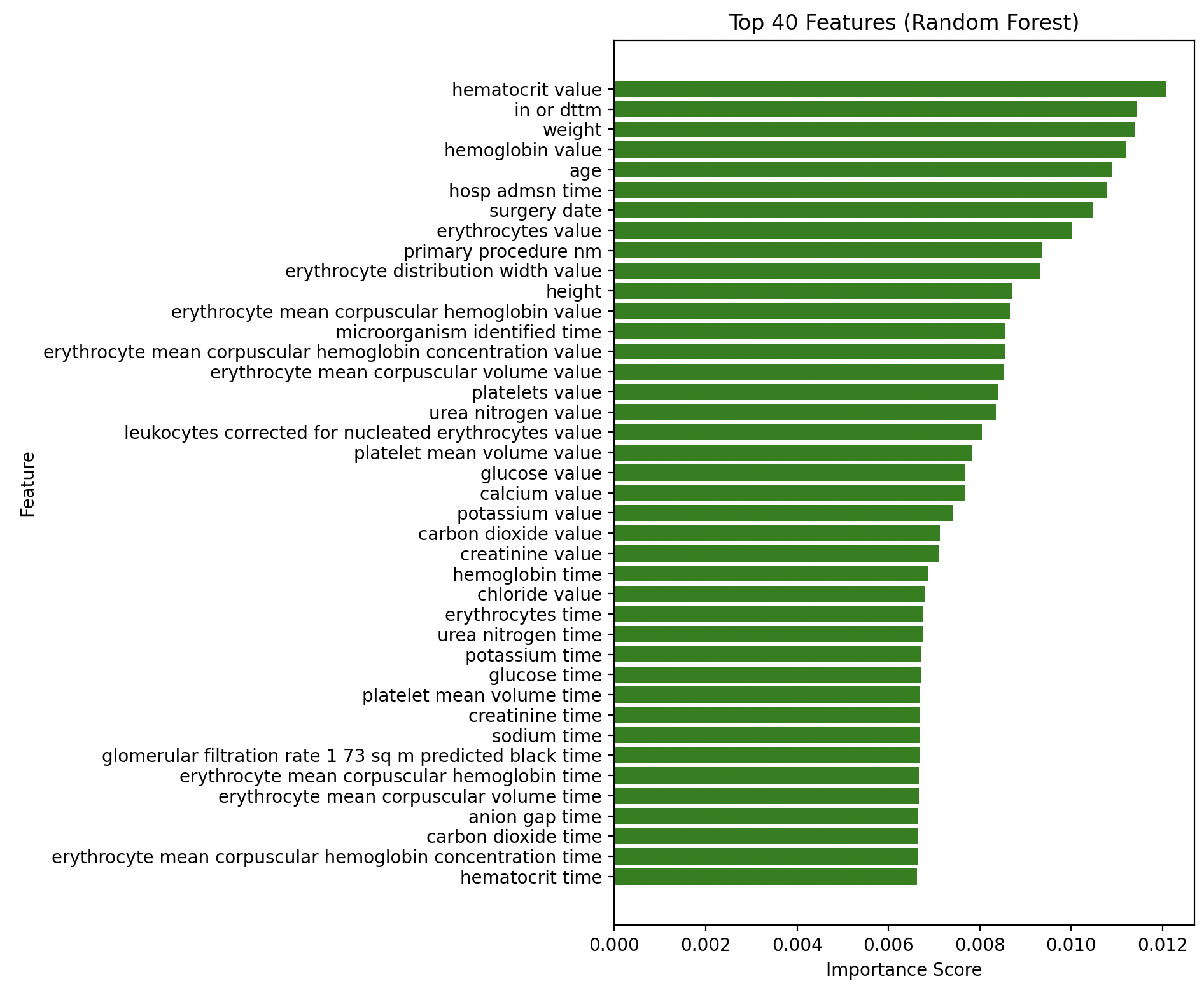}
    \caption{Gini feature importance plot for the Random Forest model, highlighting the top 40 features contributing to the model's predictions. The most influential features include 'hematocrit value,' 'operation start time (in or dtitm),' and 'weight,' followed by other significant predictors such as 'hemoglobin value,' 'age,' and 'hospital admission time.'}
    \label{fig:rf_feature_importance}
\end{figure}

\begin{figure*}[htbp]
    \centering
    % First row
    % \resizebox{1\textwidth}{!}{% scale to 90% of text width
    \begin{subfigure}[b]{0.45\textwidth}
        \includegraphics[width=\textwidth]{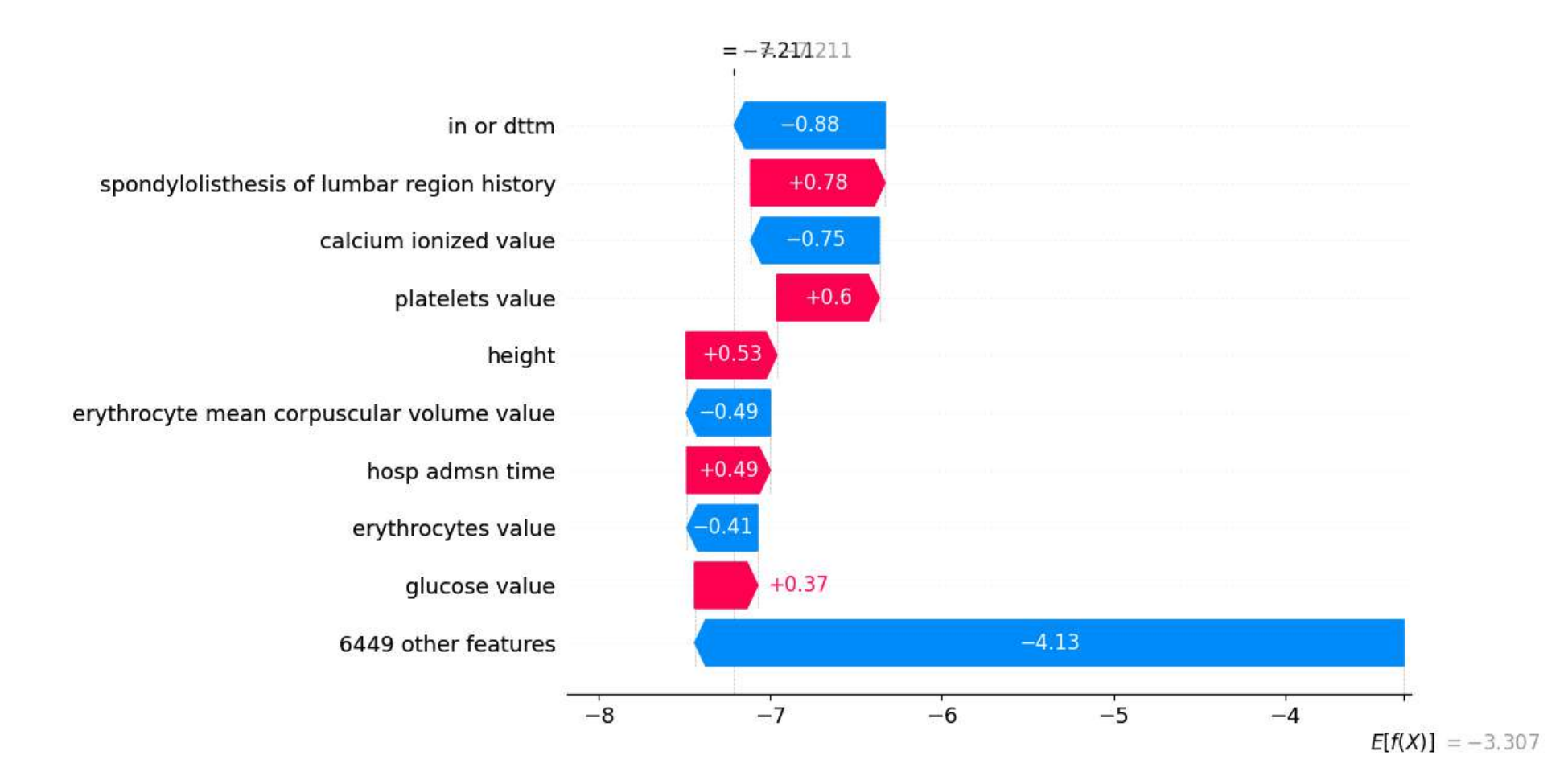}
        \caption{For Class 0, the most influential features driving predictions include operation-related attributes such as 'in or dttm' and 'spondylolisthesis of lumbar region history,' followed by laboratory measurements like 'calcium ionized value' and 'platelets value,' with 'glucose value' also contributing.}
        \label{fig:class0_waterfall}
    \end{subfigure}
    \hfill
    \begin{subfigure}[b]{0.45\textwidth}
        \includegraphics[width=\textwidth]{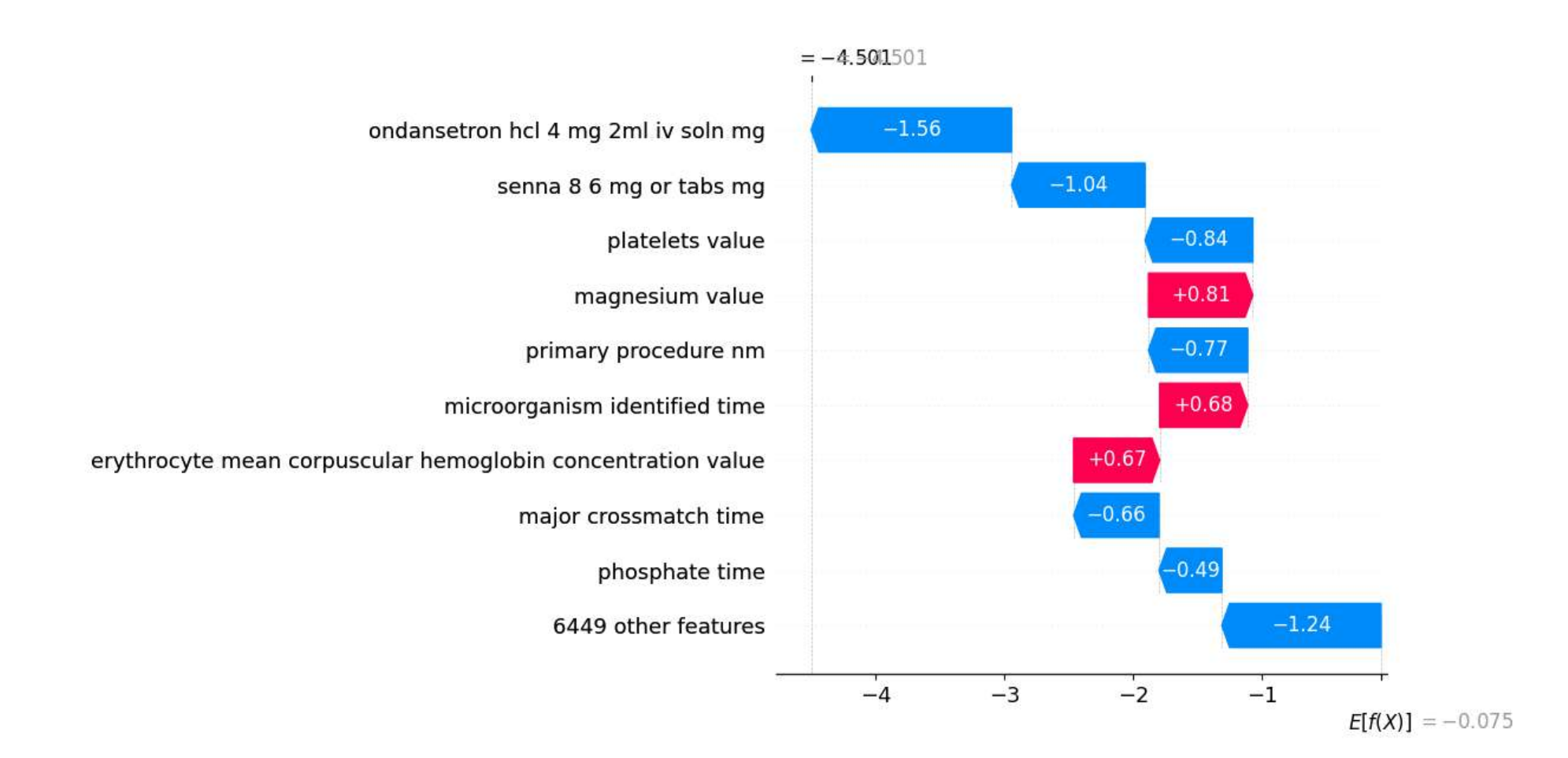}
        \caption{For Class 1, medication-related features like 'ondansetron hcl' and 'senna 8.6 mg' are significant drivers, alongside 'platelets value' and 'major crossmatch time,' indicating the importance of lab and medication data. }
        \label{fig:class1_waterfall}
    \end{subfigure}

    % Second row
    \vspace{1em}
    \begin{subfigure}[b]{0.45\textwidth}
        \includegraphics[width=\textwidth]{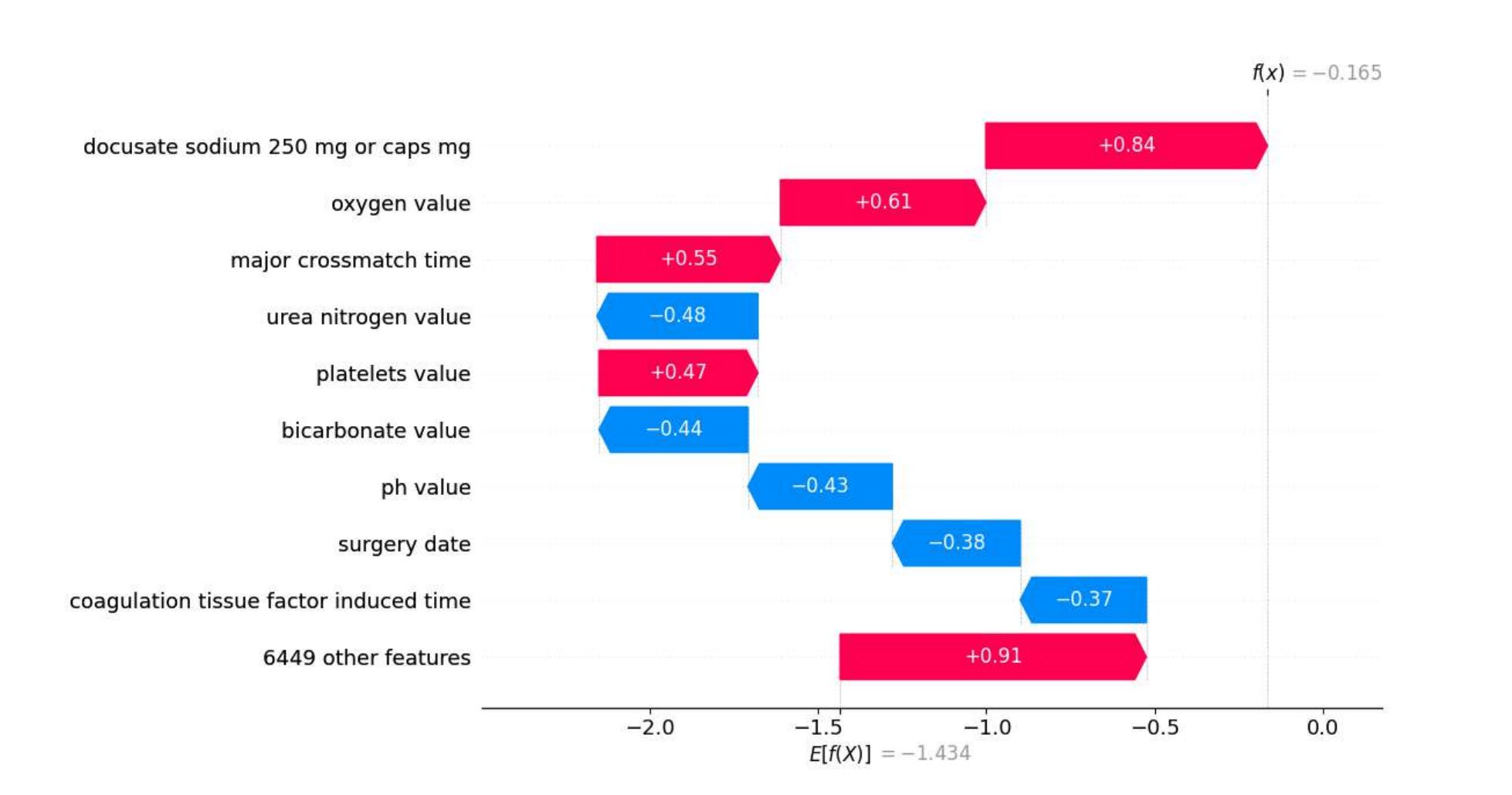}
        \caption{For Class 2, clinical measurements such as 'oxygen value' and 'bicarbonate value' along with lab features like 'platelets value' and 'urea nitrogen value,' significantly influence predictions, showing a mix of lab and vital sign contributions. }
        \label{fig:class2_waterfall}
    \end{subfigure}
    \hfill
    \begin{subfigure}[b]{0.45\textwidth}
        \includegraphics[width=\textwidth]{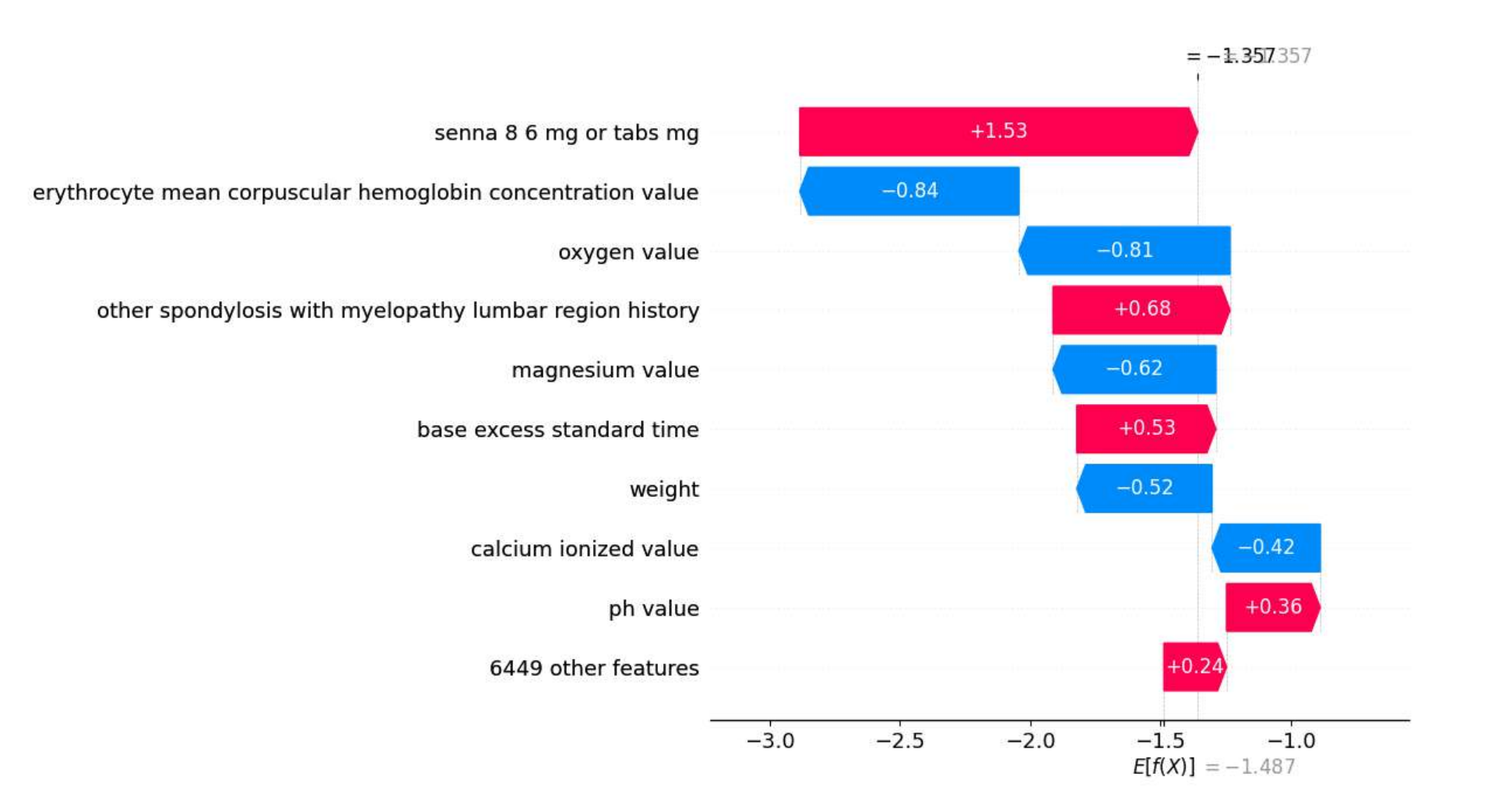}
        \caption{For Class 3, patient-specific features such as 'senna 8.6 mg,' 'erythrocyte mean corpuscular hemoglobin concentration value,' and 'calcium ionized value' dominate, while 'weight' and 'ph value' also play a role, showing the contribution of medication and vital statistics.}
        \label{fig:class3_waterfall}
    \end{subfigure}
    % }
    \caption{SHAP Waterfall Plots for One-Versus-Rest Binary Classification Model Local Interpretation. Each plot highlights the feature contributions for each class, showcasing the significant features influencing the predictions for each class.}
    \label{fig:shap_waterfall}
\end{figure*}

\subsection{Model Performance}

Our goal is to predict postoperative complication severity from preoperative EHR data. Complications are categorized into 10 clinical types (e.g., respiratory, neurological, infection), with severity quantified by the total number of complications per patient. We define four severity classes: 0 (none), 1 (one), 2 (two), and 3 (three or more complications), as illustrated in Figure~\ref{fig:overview}.\\

Table~\ref{tab:performance_metrics} compares models for predicting postoperative complication severity. Except for SVM, baselines perform similarly; tree-based methods (RF, XGBoost) slightly lead in overall accuracy—76\% (73–80\%) for RF and 73\% (68–78\%) for XGBoost—and in balanced F1 (70\% for both). FAIR-MTL is competitive, with 75\% accuracy (70–79\%) and AUC = 0.86 (95\% CI: 0.81–0.89), while offering the strongest fairness, as summarized in Table~\ref{tab:fairness_metrics_combined}. Specifically, FAIR-MTL achieves the smallest demographic-parity gaps for both gender and age and the lowest equalized-odds gap for age, while remaining competitive on equalized-odds for gender. Overall, FAIR-MTL outperforms the baselines on fairness, providing the most equitable predictions among all models compared. Figure~\ref{fig:subgroup_accuracy} also shows how FAIR-MTL achieves more uniform accuracy across sex and age groups, whereas tree-based models vary more—especially across age cohorts.

To evaluate model fairness, we assess performance disparities across sensitive attributes, gender and age, using two primary metrics: demographic parity difference (DP Diff), and equalized odds difference (EO Diff). These summarized in Table~\ref{tab:fairness_metrics_combined}. Gender is categorized as male and female, while age groups include young (0-18), adult (18–35), middle-aged (36–50), and senior (51+).

Across both sensitive attributes, FAIR-MTL consistently exhibits lower disparity metrics compared to the baseline neural network. For example, under gender, FAIR-MTL achieves an average demographic parity difference of 0.055 versus 0.066 for the NN, and maintains a comparable EO Diff (0.094 vs. 0.138). Age-related disparities are more pronounced in the NN, with an EO Diff of 0.444. In contrast, FAIR-MTL significantly reduces this disparity to 0.148.

These results demonstrate that FAIR-MTL not only maintains competitive overall accuracy (Table~\ref{tab:performance_metrics}), but also improves fairness by reducing inter-group variability and aligning true positive rates across subpopulations. While Random Forest achieves strong fairness metrics on gender (e.g., EO Diff = 0.087), its age-related fairness lags behind, with an EO Diff of 0.380, notably higher than FAIR-MTL’s 0.148. This indicates that FAIR-MTL provides a more balanced trade-off between performance and fairness across both sensitive attributes, supporting its use in equitable surgical decision support systems.

\subsection{Explainable AI (XAI)}

To enhance transparency and interpretability, we employed XAI techniques, such as SHAP and tree-based feature importance, to analyze the contributions of individual features to the model’s predictions on random forest model result. Figure \ref{fig:shap_waterfall} revealed key feature drivers for each postoperative complication severity class. For Class 0, features like "in or dttm" (negative contribution) and "spondylolisthesis of lumbar region history" (positive contribution) had significant impacts, along with lab values such as "calcium ionized value" and "platelets value". In Class 1, medication-related variables, such as "ondansetron hcl" (negative) and "magnesium value" (positive), along with clinical timings, strongly influenced predictions. For Class 2, coagulation-related features like "coagulation tissue factor induced time" (positive) and metabolic markers such as "urea nitrogen value" and "ph value" (negative) were prominent contributors. In Class 3, medication usage (e.g., "senna 8 mg", positive) and lab values (e.g., "oxygen value", negative) played pivotal roles. 
These insights demonstrate the interpretability and clinical relevance of the model’s predictions, providing valuable guidance for understanding and mitigating postoperative risks.

Figure \ref{fig:rf_feature_importance} highlights the top 40 predictors as identified by the Random Forest model using GINI importance. Among these, "hematocrit value", "in or dttm", and "weight" were the most influential, underscoring the critical role of hematological parameters, patient-specific demographics, and temporal metrics. Lab values such as "hemoglobin value", "platelets value", and metabolic markers like "calcium value" and "glucose value" also emerged as significant contributors, reflecting the importance of physiological stability in predicting postoperative complications.

\begin{table}[ht]
\centering
\caption{Overall performance and subgroup accuracy metrics for predicting postoperative mortality rate. Performance is reported using Accuracy, AUC, Precision, Recall, and F1-score. Subgroup accuracies are reported for sensitive attributes: \textbf{Sex} (Female, Male) and \textbf{Age Group} (Group 0: 0-34 years, Group 1: 35–54 years, Group 2: 55–74 years, Group 3: 75+ years). Subgroup accuracy represents the average accuracy across four complication severity classes, where smaller differences between subgroups indicate more equitable model behavior.}
\resizebox{1 \linewidth}{!}{
\begin{tabular}{lccccc}
\toprule
\textbf{Metric} & \textbf{Random Forest} & \textbf{XGBoost} & \textbf{SVM} & \textbf{Neural Network} & \textbf{FAIR-MTL} \\
\midrule
\multicolumn{6}{l}{\textbf{Overall Performance}} \\
\midrule
Accuracy (CI) & \makecell{87.17\% \\ (85.92\%–88.32\%)} 
              & \makecell{87.90\% \\ (86.68\%–89.02\%)} 
              & \makecell{88.33\% \\ (85.61\%–88.03\%)} 
              & \makecell{86.87\% \\ (85.61\%–88.03\%)} 
              & \makecell{89.89\% \\ (88.41\%–91.20\%)} \\
AUC (CI) & \makecell{86.04\% \\ (80.25\%–84.53\%)} 
         & \makecell{86.04\% \\ (84.52\%–88.22\%)} 
         & \makecell{86.35\% \\ (84.19\%–87.71\%)} 
         & \makecell{86.89\% \\ (85.08\%–88.53\%)} 
         & \makecell{89.01\% \\ (86.78\%–91.13\%)} \\
Precision               & 80.15\%                    & 78.43\%                    & 86.36\%                    & 84.64\%                    & 79.85\%                    \\
Recall                  & 71.58\%                    & 67.18\%                    & 67.21\%                    & 77.21\%                    & 79.09\%                    \\
F1-score                & 74.71\%                    & 70.58\%                    & 71.81\%                    & 80.21\%                    & 79.33\%                    \\
\specialrule{\heavyrulewidth}{0pt}{0pt}

\multicolumn{6}{l}{\textbf{Subgroup Accuracy \& AUC by Sex}} \\
\midrule
Female – Accuracy       & 90.08\% & 90.08\% & 90.44\% & 87.62\% & 88.96\% \\
Female – AUC            & 84.06\% & 84.06\% & 83.36\% & 84.45\% & 88.98\% \\
Male – Accuracy         & 83.53\% & 85.18\% & 85.70\% & 81.81\% & 91.07\% \\
Male – AUC              & 87.06\% & 87.06\% & 87.73\% & 88.15\% & 89.05\% \\

\midrule
\multicolumn{6}{l}{\textbf{Subgroup Accuracy \& AUC by Age Group}} \\
\midrule
G0 – Accuracy & 95.96\% & 95.96\% & 96.30\% & 96.30\% & 90.59\% \\
G0 – AUC      & 84.62\% & 84.62\% & 87.57\% & 87.34\% & 89.59\% \\
G1 – Accuracy& 92.27\% & 92.15\% & 93.24\% & 92.27\% & 91.30\% \\
G1 – AUC     & 86.15\% & 86.15\% & 80.21\% & 85.60\% & 89.46\% \\
G2 – Accuracy& 86.61\% & 86.90\% & 87.77\% & 86.33\% & 89.67\% \\
G2 – AUC     & 83.38\% & 83.38\% & 84.84\% & 84.67\% & 89.38\% \\
G3 – Accuracy  & 74.85\% & 78.70\% & 76.88\% & 70.79\% & 87.76\% \\
G3 – AUC       & 78.66\% & 78.66\% & 82.28\% & 81.15\% & 86.74\% \\
\bottomrule
\end{tabular}}
\label{tab:inspire-performance}
\end{table}

\begin{table}[htbp]
\centering
\caption{
Fairness metrics—Demographic Parity (DP) and Equalized Odds (EO) differences—for gender and age subgroups across all models (lower is better).  
FAIR-MTL achieves the lowest disparities on both metrics and sensitive attributes, indicating the strongest fairness performance.
}
\resizebox{0.85\linewidth}{!}{
\begin{tabular}{lcccc}
\hline
\textbf{Model} & 
\makecell{\textbf{DP Diff} \\ \textbf{(Gender)}} & 
\makecell{\textbf{DP Diff} \\ \textbf{(Age)}} & 
\makecell{\textbf{EO Diff} \\ \textbf{(Gender)}} & 
\makecell{\textbf{EO Diff} \\ \textbf{(Age)}} \\
\hline
FAIR-MTL & 0.00618 & 0.01437 & 0.03388 & 0.05404 \\
XGBoost  & 0.01678 & 0.12641 & 0.04208 & 0.24255 \\
RF       & 0.02576 & 0.21052 & 0.04266 & 0.33553 \\
SVM      & 0.03657 & 0.14173 & 0.02896 & 0.24182 \\
NN       & 0.07985 & 0.28103 & 0.06570 & 0.19521 \\
\hline
\label{tab:inspire-fairness}
\end{tabular}
}

\end{table}

\subsection{External Validation Study on INSPIRE Dataset}

To evaluate the generalization of our approach, we used a secondary dataset, INSPIRE, to compare FAIR-MTL with all baseline models. We formulated the predictive modeling task as a binary classification problem, with the target variable indicating postoperative mortality (0 = survived, 1 = deceased). However, the dataset presents significant biases. The outcome distribution is imbalanced, with the majority class accounting for 86.03\% of all samples. In addition, demographic disparities are evident. The male-to-female ratio is approximately 5.3:1, indicating strong overrepresentation of male patients. Age distribution is also skewed: Group 0 (0–34 years) represents about 2.6\% of the cohort, Group 1 (35–54 years) about 8.2\%, Group 2 (55–74 years) dominates with 41.5\%, and Group 3 (75+ years) makes up 14.5\%. These imbalances highlight the need to carefully consider data fairness, as models trained on such distributions may generalize poorly across underrepresented subgroups.

Across all models, FAIR-MTL achieves the highest overall accuracy at 89.89\% (95\% CI: 88.41--91.20) and shows the most consistent performance across demographic subgroups, as shown in Table \ref{tab:inspire-performance}. For gender, the accuracy gap between females (88.96\%) and males (91.07\%) is only 2.11 percentage points, compared to 4.90 for XGBoost and larger gaps for other models. For age groups, FAIR-MTL’s maximum gap is 1.44 points, versus 12.64 for XGBoost and 21.05 for Random Forest. FAIR-MTL also performs better under fairness metrics, as shown in Table \ref{tab:inspire-fairness}, with the lowest demographic parity differences (0.00618 for sex, 0.01437 for age) and equalized odds differences (0.03388 for sex, 0.05404 for age). These results indicate that FAIR-MTL delivers strong predictive accuracy while minimizing disparities across gender and age.

Paired t-test analyses with bootstrap resampling confirmed that the reductions in fairness disparities achieved by FAIR-MTL are not only large in magnitude but also highly statistically significant. Across all comparisons with baseline models (NN, SVM, RF, XGB), the t-statistics were strongly negative (indicating consistently lower disparities for FAIR-MTL), and the $p$-values were vanishingly small (all $p<10^{-36}$), leaving negligible probability that the observed improvements are attributable to random variation. Comparable levels of statistical robustness were observed across all subgroups and metrics, underscoring that FAIR-MTL achieves consistently lower fairness gaps with strong statistical significance.

\subsection{Ablation Study on FAIR-MTL} 
Table~\ref{tab:ablation} presents an ablation study evaluating the contribution of key components in the FAIR-MTL architecture, including reweighting, shared layers, and task-specific heads. The results demonstrate that removing any of these components leads to a noticeable degradation in either accuracy or fairness metrics, reflected by higher demographic parity (DP) and equalized odds (EO) differences across both age and gender groups. While a standard neural network without multi-task heads achieves higher accuracy (0.77) and AUC (0.89), it exhibits the worst fairness performance, with the largest DP and EO gaps. In contrast, the full FAIR-MTL model maintains competitive predictive performance (Accuracy: 0.75, AUC: 0.85) while significantly reducing fairness disparities. These findings highlight the necessity of each design component for achieving a balanced trade-off between accuracy and fairness.
Overall, these results suggest that our multitask decomposition approach not only maintains competitive predictive performance but also leads to more equitable outcomes across demographic groups.

\subsection{Discussion}

\begin{table}[]
\centering
\caption{
FAIR-MTL Ablation study.  
Each model variant omits one component, reweighting, shared layers, or task-specific heads, to assess its impact on predictive performance and fairness.  
Excluding any component increases demographic-parity (DP) and equalized-odds (EO) disparities across sensitive attributes, confirming the contribution of every module.
}

\label{tab:ablation}
\fontsize{9}{11}\selectfont
\resizebox{\columnwidth}{!}{
\begin{tabular}{lcccccc}
\toprule
\textbf{Model} & 
\textbf{Accuracy} & 
\textbf{AUC} & 
\makecell{\textbf{DP Diff} \\ \textbf{(Age)}} & 
\makecell{\textbf{EO Diff} \\ \textbf{(Age)}} & 
\makecell{\textbf{DP Diff} \\ \textbf{(Gender)}} & 
\makecell{\textbf{EO Diff} \\ \textbf{(Gender)}} \\
\midrule
FAIR-MTL & 0.75 & 0.85 & 0.056 & 0.148 & 0.055 & 0.094 \\
FAIR-MTL (no reweighting) & 0.70 & 0.81 & 0.089 & 0.397 & 0.012 & 0.054 \\
FAIR-MTL (no shared layers) & 0.72 & 0.80 & 0.0291 & 0.0937 & 0.0291 & 0.0937 \\
NN (no task heads) & 0.77 & 0.89 & 0.145 & 0.444 &0.066 & 0.138\\

\bottomrule
\end{tabular}}
\end{table}
We developed and evaluated a range of machine learning models to predict postoperative complication severity in patients undergoing spinal fusion surgery. Among traditional classifiers, Random Forest (RF) and XGBoost achieved strong predictive performance, with accuracies of 76\% and 73\% and AUCs of 0.79 each. A custom neural network model further improved performance, achieving an AUC of 0.89, highlighting the utility of deep learning in capturing complex, non-linear relationships in clinical data.

To address concerns of model fairness and equity, we introduced a fairness-aware multitask learning (FAIR-MTL) framework that decomposes the prediction task based on inferred sensitive subgroups. While FAIR-MTL showed slightly lower overall AUC of 0.86 compared to the best performing model, it still exceeds all the other models in AUC performance while significantly reduces disparity metrics across gender and age groups, achieving lower demographic parity and equalized odds gaps without substantial loss in accuracy. These results demonstrate the potential of task decomposition to balance predictive power with subgroup equity, offering a more reliable foundation for decision support in high-stakes clinical contexts.

We also conducted model interpretation using both Gini importance and SHAP values on Random Forest. These analyses consistently identified key predictors such as hematocrit, operation start time, weight, and other laboratory values. SHAP-based explanations provided both global and patient-specific insights, enabling clinicians to identify modifiable risk factors during preoperative assessment. For instance, flagged abnormalities in hematocrit or body weight could prompt early intervention, potentially improving surgical outcomes and patient recovery trajectories.

A core contribution of this study lies in its fairness-aware modeling strategy. Standard neural networks showed meaningful disparities across age groups, with equalized odds gaps exceeding 0.44 in some cases. In contrast, FAIR-MTL achieved more consistent group-wise accuracy and substantially reduced disparity metrics (e.g., EO gap: 0.148 for age vs. 0.444 in NN). Moreover, it maintained high minimum subgroup F1 scores and lower standard deviation across groups, reinforcing its robustness across diverse patient populations. As clinical AI becomes more embedded in care delivery, such approaches are essential to ensure that algorithmic tools do not inadvertently exacerbate healthcare inequities.

\section{Conclusion}
Rather than relying on a single model trained on the entire cohort—where the statistical dominance of majority groups can overshadow minority populations—FAIR-MTL adopts a subgroup-aware strategy. By clustering patients into demographic subgroups with intrinsically similar characteristics and routing them to submodels trained specifically for those subgroups, FAIR-MTL ensures that the learning dynamics of minority groups are not overwhelmed by majority patterns, thereby promoting fairness in model outcomes. At the same time, FAIR-MTL enables knowledge sharing across subgroups, striking a balance between specialization and generalization. By jointly achieving fairness, interpretability, and accuracy, FAIR-MTL not only advances predictive modeling for spinal surgery but also lays the groundwork for ethically aligned AI systems that deliver equitable predictions in clinical decision support.

\begin{acks}
This research was supported by Shriners Children's Hospital and the Georgia Institute of Technology through the Forecasting Unexpected Signal Change in Posterior Spinal Fusion (FUSION) Project. Additional support was provided in part by the AI Makerspace of the College of Engineering and other research cyberinfrastructure resources and services offered by the Partnership for an Advanced Computing Environment (PACE) at the Georgia Institute of Technology, Atlanta, Georgia, USA.
We also gratefully acknowledge Wallace H. Coulter Distinguished Faculty Fellowship, a Petit Institute Faculty Fellowship, and research funding from Amazon and Microsoft Research awarded to Professor May D. Wang.
\end{acks}

\bibliographystyle{ACM-Reference-Format}
\bibliography{ref}
% \newpage

\end{document}